\documentclass[12pt]{article}

\usepackage{latexsym}

\long\def\nop#1{}

\def\comment{\edef\cps{\the\parskip} \parskip=0.5cm \begingroup \tt}

\expandafter\ifx\csname shortcite\endcsname\relax
\let\shortcite=\cite
\fi

\newbox\current

\long\def\plframebox#1{
\setbox\current\vbox{#1}		

\vbox to \ht\current {\hrule\vss
\hbox to \wd\current {%
\vrule \hss\box\current\hss \vrule}
\vss\hrule }
}

\long\def\eatpar#1{%
\ifx#1\par                      
\let\nextmove=\eatpar           
\else
\let\nextmove=#1
\fi
\noexpand\nextmove
}

\def\modifymargins#1#2{
\newdimen\addtoh
\newdimen\addtow
\addtoh=#1
\addtow=#2

\advance\topmargin by -\addtoh
\multiply\addtoh by 2
\advance\textheight by \addtoh

\advance\oddsidemargin by -\addtow
\advance\evensidemargin by -\addtow
\multiply\addtow by 2
\advance\textwidth by \addtow
}

\begingroup
\catcode`\~=11
\gdef\centertilde#1{\lower #1pt\hbox{~}}
\endgroup

\newcount\currenttime
\newcount\hour
\newcount\minute

\def\printtime{%
\currenttime=\time
\hour=\currenttime
\divide\hour by 60
\minute=-\hour
\multiply\minute by 60
\advance\minute by \currenttime
\the\hour:\ifnum\minute<10 0\fi\the\minute
}

\begingroup
\makeatletter
\global\let\@@date=\@date
\gdef\@date{\@@date\ --- \printtime}
\endgroup

\def\oggi{\number\day\space 
\ifcase\month\or
Gennaio\or Febbraio\or Marzo\or Aprile\or Maggio\or Giugno\or
Luglio\or Agosto\or Settembre\or Ottobre\or Novembre\or Dicembre\fi
\space \number\year}

\newcounter{rmexample}

\def\proof{\noindent {\sl Proof.\ \ }}

\def\qed{\hfill{\boxit{}}
  \ifdim\lastskip<\medskipamount \removelastskip\penalty55\medskip\fi}
\def\qedn#1{\hfill{\boxit{}$_#1$}
  \ifdim\lastskip<\medskipamount \removelastskip\penalty55\medskip\fi}
\long\def\boxit#1{\vbox{\hrule\hbox{\vrule\kern3pt
                  \vbox{\kern3pt#1\kern3pt}\kern3pt\vrule}\hrule}}

  \def\D{{\cal D}}

\def\ie{i.e.}
\def\eg{e.g.}
\def\wrt{w.r.t.}

\def\l{\langle}
\def\r{\rangle}

\def\true{{\sf true}}
\def\false{{\sf false}}

\def\np{{\rm NP}}
\def\conp{{\rm coNP}}
\def\Dp{${\rm D}^p$}
\def\S#1{\mbox{$\Sigma^p_{#1}$}}
\def\P#1{\mbox{$\Pi^p_{#1}$}}
\def\D#1{\mbox{$\Delta^p_{#1}$}}

\def\nuc#1{\mbox{$\parallel\!\leadsto$#1}}

\def\nucS#1{\nuc{$\Sigma^p_{#1}$}}
\def\nucP#1{\nuc{$\Pi^p_{#1}$}}
\def\nucD#1{\nuc{\D{#1}}}

\def\profont{\sf}

\def\sat{{\profont sat}}
\def\unsat{{\profont unsat}}

\def\x3c{{\profont x3c}}

\def\possnewtheorem#1#2{
\expandafter\ifx\csname #1\endcsname\relax
\newtheorem{#1}{#2}
\fi
}

\def\possnewtheoremthree#1[#2]#3{
\expandafter\ifx\csname #1\endcsname\relax
\newtheorem{#1}[#2]{#3}
\fi
}

\possnewtheorem{theorem}{Theorem}
\possnewtheorem{corollary}{Corollary}
\possnewtheorem{lemma}{Lemma}
\possnewtheoremthree{proposition}[theorem]{Proposition}
\possnewtheorem{definition}{Definition}
\possnewtheorem{question}{Question}
\possnewtheorem{example}{Example}
\possnewtheorem{nontheorem}{Counterexample}
\possnewtheorem{property}{Property}
\possnewtheorem{assumption}{Assumption}
\possnewtheorem{conjecture}{Conjecture}
\possnewtheorem{notation}{Notation}
\newenvironment{theorem*}[1]{{\noindent \bf Theorem~#1}\begin{it}}{\end{it}\

}

\def\rr#1{*_{RR(#1)}}

\def\prec{prec}
\def\cons{cons}
\def\just{just}

\def\ll{\left\langle}
\def\rr{\right\rangle}

\title{Where Fail-Safe Default Logics Fail}
\author{Paolo Liberatore%
\thanks{Dipartimento di Informatica e Sistemistica,
Universit\`a di Roma ``La Sapienza'',
Via Salaria 113, 00198, Roma, Italy.
Email: {\tt paolo@liberatore.org}}}
\date{}

\begin{document}

\maketitle

\begin{abstract}

Reiter's original definition of default logic allows for the
application of a default that contradicts a previously
applied one. We call {\em failure} this condition. The
possibility of generating failures has been in the past
considered as a {\em semantical} problem, and variants have
been proposed to solve it. We show that it is instead a {\em
computational} feature that is needed to encode some domains
into default logic.

\end{abstract}
 %

\begin{verbatim}

\end{verbatim}

\newpage\tableofcontents\newpage

\section{Introduction}

Since the introduction of default logic \cite{reit-80},
semantical problems of the original definition have been
identified, and variants have been proposed to solve them
\cite{luka-88,brew-91-b,rych-91,delg-scha-jack-94,gior-mart-94,miki-trus-95}.
One of the problems with Reiter's definition is that the
application of a sequence of defaults may lead to failure.
The following example shows this problem.

\[
T = \ll \left\{\frac{a:b}{c}, \frac{c:a}{\neg b}\right\}, \{a\} \rr
\]

The default $\frac{a:b}{c}$ entails $c$ whenever $a$ is true
and $b$ is consistent with our current knowledge. The
application of this default makes $c$ true, therefore making
the second default $\frac{c:a}{\neg b}$ applicable, which
makes $b$ false. This result is however in contradiction
with the assumption of consistency of $b$ we made for
applying the first default.

We call {\em failure} the condition in which the application
of another default makes a default that has already been
applied inapplicable. In other words, if the application of
a default contradicts the assumption of a default that has
already been applied, this is a failure. The possibility of
failures has been in the past considered a drawback,
especially because failures may make the evaluation of
theories like $T$ impossible (\ie, these theories have no
extensions.)

A typical solution to this problem is to refrain from
applying defaults that would lead to failure. This is done
by justified default logic~\cite{luka-88}, constrained
default logic \cite{delg-scha-jack-94}, and cumulative
default logic \cite{brew-91-b,gior-mart-94}. For the theory
$T$ above, the application of the first default makes $c$
true, but the second default is not applied because it would
lead to failure. We therefore conclude that $T$ is
equivalent to the propositional theory $\{a,c\}$.

This solution is {\em semantically} good, as it it allows
for the evaluation of theories like $T$ (\ie, it assigns
extensions to theories that would otherwise have none.) On
the other hand, having forbidden failures can be seen as a
{\em computational} problem, as the encoding of some domains
requires exactly this ``ability to fail.'' An example is the
translation from reasoning about actions to default logic
proposed by Turner~\cite{turn-97}. This translation
generates all possible evaluations of a variable $x$ by
means of a pair of defaults $\frac{:x}{x}$ and $\frac{\neg
x}{\neg x}$, and then removes the unwanted evaluations by
generating failures. For example, the simple theory
$\{Holds(Alive,S_0)\}$, telling that Fred is alive in the
initial state, is translated into the following default
theory:

\[
\ll \left\{
\frac{:Holds(Alive,S_0)}{Holds(Alive,S_0)} ,~
\frac{:\neg Holds(Alive,S_0)}{\neg Holds(Alive,S_0)} ,~ 
\frac{\neg Holds(Alive,S_0):}{\false} \right\} ,~
\emptyset
\rr
\]

Either the first or the second default can be applied, but
not both. Depending on which one we decide to apply, we
obtain $Holds(Alive,S_0)$ or $\neg Holds(Alive,S_0)$. The
third default generates a failure whenever $\neg
Holds(Alive,S_0)$ is true. The only possible remaining case
is therefore that in which $Holds(Alive,S_0)$ is true. In
other words, the first two defaults generate all possible
evaluations of $Holds(Alive,S_0)$ and the third default
deletes the one we do not want. In general, some defaults
generate all possible evaluations of the fluents and some
other defaults delete the ones that are not possible. The
latter are called {\em killing defaults} by Cholewinski,
Marek, Mikitiuk, and Truszczynski \shortcite{chol-etal-95},
who presented other reductions where killing defaults are
used to delete unwanted solutions. Since this deletion is
realized by generating a failure, these translations do not
work for semantics where failure is impossible, such as
justified default logic \cite{luka-88}. The inability to
fail can be therefore seen as a limitation, as some
translations from other formalisms into default logics
require failures.

In this paper, we consider the problem of translating
default semantics that can fail (fail-prone) into default
semantics where failure is impossible (fail-safe). In order
for the results to abstract over the specific semantics, we
consider a sufficiently general definition of ``semantics of
default logic'', based on the concept of process
\cite{froi-meng-92,anto-sper-94,froi-meng-94,anto-99}. We
formalize the concept of fail-safeness in this framework,
and investigate the translability from fail-prone into
fail-safe semantics.

The translations we consider are polynomial either in time
or size of the result. We consider translations that
preserve the skeptical consequences, or the extensions, or
the processes of a default theory.

The least constrained form of translation is that
translating the inference problem: given a theory $\l D,W
\r$ and a formula $p$ we require the translation to produce
another theory $\l D',W' \r$ and another formula $p'$ in
such a way $\l D,W \r \models p$ holds in one semantics if
and only if $\l D',W' \r \models p'$ holds in the other one.
Such translations are possible in polynomial time between
all semantics that have the same complexity. For example, we
can translate Reiter's default logic into justified default
logic in this way as both semantics are \P{2}-complete
\cite{gott-92-b,stil-92-b,cado-scha-93}.

We extend this result by simplifying the translation of $p$
into $p'$: we indeed show a translation such that $\l D,W \r
\models p$ holds if and only if $\l D',W' \r \models a \vee
p$ holds, where $a$ is a new variable. Intuitively, the new
variable expresses the condition of failure in a semantics
that cannot fail: by setting $a$ to true whenever a failure
should be necessary, the generated extension implies $a \vee
p$, and is thus irrelevant to skeptical entailment.

Using this translation, the extensions of the theories $\l
D,W \r$ and $\l D',W' \r$ are not the same. We therefore
consider translations that preserve the extensions: such
translations are called faithful \cite{kono-88,gott-95}.
Clearly, no faithful translation is possible from a
semantics that may not have extensions to one that always
have. This is how Delgrande and Schaub \cite{delg-scha-03},
for example, have shown that Reiter's default logic cannot
be translated into justified default logic. However, the
question remains if this impossibility is only due to a
possible lack of extensions. We therefore restrict to the
case in which the original theory has at least one extension
and prove that an extension-preserving translation exists or
not depending on what we assume to be polynomial: the
running time of the translation or the size of the result.
 
Finally, we consider translations preserving the processes
of a default theory. Processes are the basic semantic notion
of the operational semantics for default logic
\cite{froi-meng-92,anto-sper-94,froi-meng-94,anto-99}; they
are sequences of defaults that can be applied in a default
theory. Since two or more processes may correspond to the
same extension, two default theories may have the same
extensions but different processes. Therefore, translations
that preserve the processes can be considered as the ``most
preserving'' ones.

 %

\section{Definitions}

\subsection{Default Logics}

We use the operational semantics for default logics. Two
slightly different operational semantics for default logics
have been given independently by Antoniou and Sperschneider
\cite{anto-sper-94,anto-99} and by Froidevaux and
Mengin~\cite{froi-meng-92,froi-meng-94}. A default is a rule
of the form:

\[
d=\frac{\alpha:\beta}{\gamma}
\]

The formulae $\alpha$, $\beta$, and $\gamma$ are called the
precondition, the justification, and the consequence of $d$,
and are denoted as $\prec(d)$, $\just(d)$, and $\cons(d)$,
respectively. This notation is extended to sets and
sequences of defaults in the obvious way. A default is
applicable if its precondition is true and its justification
is consistent; if this is the case, its consequence should
be considered true.

A default theory is a pair $\l D,W \r$ where $D$ is a set of
defaults and $W$ is a consistent theory, called the
background theory. We assume that all formulae are
propositional, the alphabet and the set $D$ are finite, and
all defaults have a single justification. The assumption
that $W$ is consistent is not standard; however, all known
semantics give the same evaluation when the background
theory is inconsistent. We define a process to be a sequence
of defaults that can be applied starting from the background
theory.

\begin{definition}

A {\em process} of a default theory $\l D,W \r$ is a
sequence of defaults $\Pi$ such that $W \cup \cons(\Pi)$ is
consistent and $W \cup \cons(\Pi[d]) \models \prec(d)$ for
every default $d \in \Pi$, where $\Pi[d]$ denotes the
sequence of defaults preceeding $d$ in $\Pi$.

\end{definition}

The definition of processes only takes into account the
preconditions and the consequences of defaults. This is
because the interpretation of the justifications depends on
the semantics. All semantics select a set of processes that
satisfy two conditions: successfulness and closure.
Intuitively, successfulness means that the justifications of
the applied defaults are not contradicted; closure means
that no other default should be applied.

The particular definition of successfulness and closure
depend on the specific semantics. The following are the
definitions used by Reiter's and constrained default logic.

\begin{description}

\item[Successfulness:] \

\begin{description}

\item[Local:] for each $d \in \Pi$, the set
$W \cup \cons(\Pi) \cup \just(d)$ is consistent;

\item[Global:]
$W \cup \cons(\Pi) \cup \just(\Pi)$ is consistent.

\end{description}

\item[Closure:]

\

\begin{description}

\item[Inapplicability:] no default $d \not\in \Pi$
is applicable to $\Pi$, \ie, either
$W \cup \cons(\Pi) \not\models \prec(d)$ or
$W \cup \cons(\Pi) \cup \just(d)$ is inconsistent;

\item[Maximality:] for any $d \not\in \Pi$, the sequence
$\Pi \cdot [d]$ is not a globally successful process.

\end{description}

\end{description}

We abstract the notions of successfulness and closure from
the particular semantics, and define them to be two
conditions on $\Pi$ and $\l D,W \r$ such that:

\begin{enumerate}

\item successfulness is antimonotonic: if $\Pi \cdot \Pi'$
is successful, then $\Pi$ is successful;

\item $[~]$ is a successful process, if $W$ is consistent;

\item successfulness and closure can be expressed as the
combination of a number of consistency tests over the
formulae in $\Pi$ and $\l D,W \r$;

\item these consistency tests are independent on the order
of defaults in $\Pi$;

\item the combination of the results of these consistency
tests results can be done in polynomial time.

\end{enumerate}

A default logic semantics defined in terms of two conditions
of successfulness and closure that satisfy these assumptions
is called {\em regular}. Most default logic semantics are
regular: the only exception known to the author is the
semantics of concise extensions, which is not regular
because the condition of subsumption requires checking all
possible default orderings \cite{rych-91}.

We remark that the last condition does not imply that
successfulness and closure can be checked in polynomial
time: they can only be checked in polynomial time once the
consistency/entailment tests have been done. This is
typically the case: for example, Reiter's default logic
closure condition amounts to check whether, for every $d
\not\in \Pi$, either $W \cup \cons(\Pi) \not\models
\prec(d)$ or $W \cup \cons(\Pi) \cup \just(d)$ is
inconsistent. Once we have checked the consistency of $W
\cup \cons(\Pi) \cup \neg \prec(d)$ and $W \cup \cons(\Pi)
\cup \just(d)$ for every $d \not\in \Pi$, determining
whether closure is satisfied can be done in linear time.

An {\em extension} of a default theory is the deductive
closure of $W \cup \cons(\Pi)$ where $\Pi$ is a successful
and closed process. Note that more than one process may
generate the same extension. The {\em skeptical}
consequences of a default theory are the formulae that are
entailed by all its extensions. The {\em credulous}
consequences are those implied by some of its extensions.

Fail-safeness of a semantics is formalized as follows.

\begin{definition}[Fail-Safe Semantics]

A regular semantics for default logic is fail-safe if, for
every default theory, any successful process is the prefix
of a successful and closed process.

\end{definition}

This definition formalizes the idea that a sequence of
defaults cannot generate a failure: if we can apply a
sequence of defaults, then an extension will be eventually
generated, possibly after applying some other defaults. In
other words, the situation in which we apply some defaults
but then find out that we do not generate an extension never
occurs. Fail-safeness is a form of commitment to defaults:
if we apply a default, we never end up with contradicting
its assumption.

Fail-safeness can also be seen as a form of monotonicity of
processes \wrt\  to sets of defaults: if a semantics is
fail-safe, then adding some defaults to a theory may only
extend the successful and closed process of the theory and
create new ones. However, this form of monotonicity is not
the same as that typically used in the literature, which is
defined in terms of consequences, not processes. Froidevaux
and Mengin~\cite[Theorem~29]{froi-meng-94} have proved a
result that essentially states that every semantics in which
closure is defined as maximal successfulness is fail-safe.
As a result, justified and constrained default logics are
fail-safe.

We recall that we assume that the background theory $W$ is
consistent. In this case, if a semantics is fail-safe, then
every default theory has a successful and closed process:
since the process $[~]$ is successful, a process $\Pi$ that
is successful and closed exists. The condition of
antimonotonicity provides an algorithm for finding this
successful and closed process: if $\Pi$ is successful and
closed, all its initial fragments are successful as well. We
can therefore obtain a successful and closed process by
iteratively adding to $[~]$ a default that lead to a
successful process.

While all fail-safe semantics give extensions to theories,
the converse is only true in some cases. According to
Reiter's and rational default logics the simple default
theory $\l \{\frac{:a}{\neg a}\},\emptyset \r$ has no
extension, and these two semantics are therefore not
fail-safe. On the other hand, every default theory has at
least one concise extension \cite{rych-91}, while the
semantics of concise extensions is not fail-safe, as shown
by the following example.

\begin{eqnarray*}
T &=& \l \{d_1,d_2\},\emptyset \r
\\
\mbox{\hbox to 0pt{where}} &&
\\
d_1 &=& \frac{:x}{x}
\\
d_2 &=& \frac{:x \wedge y}{x \wedge y}
\end{eqnarray*}

The default $d_1$ is applicable to $W=\emptyset$; the
default $d_2$ is applicable to $[d_1]$ because $d_2$ is not
subsumed by $d_1$. On the other hand, the process
$[d_1,d_2]$ is not concise, as $d_1$ is subsumed by $d_2$.
As a result, $[d_1]$ is a successful process, but is not the
initial part of any successful and closed process under the
semantics of concise extensions.

\subsection{Translations}

In this paper, we investigate the extent to which fail-safe
semantics are less expressive than fail-prone ones. In
particular, we study whether one of the two facts below can
be proved:

\begin{enumerate}

\item every theory under an arbitrary regular default
semantics can be translated into a normal default theory
under Reiter's semantics;

\item there are theories under Reiter's semantics that
cannot be translated into whichever fail-safe semantics.

\end{enumerate}

Since Reiter's default logic is fail-safe when restricting
to normal defaults, the existence of a translation of the
first kind implies that every regular default semantics can
be translated into at least one fail-safe semantics.
However, the restriction to normal defaults makes such a
result more general. Since most of the semantics for default
logic behave like Reiter's on normal defaults (an exception
is Rychlik's concise semantics \cite{rych-91}, which is
however not fail-safe), this result extends to all these
semantics. In order to simplify the terminology, we formally
define ``normal default logic'' as a semantics for default
logic.

\begin{definition}

Normal default logic is the restriction of Reiter's default
logic to the case of normal defaults.

\end{definition}

Ideally, we would like results of the first kind to be
exactly the converse of the second one, \ie, every regular
semantics can be translated into every fail-safe semantics.
This general question has however an easy (and of little
significance) negative answer: the semantics that has $[~]$
as the only successful and closed process is fail-safe, but
none of the considered semantics can be translated into it.
Indeed, consequence-preserving translations are impossible
in polynomial time because entailment is \conp-complete in
this semantics but \P{2}-complete in most semantics;
faithful translations are impossible because this semantics
always gives a single extension to a theory while the other
one may give more.

The results about the existence of translations depend on
what we require from the translations. A minimal requirement
is that the consequences are preserved. At the other
extreme, we may require a translation to preserve the set of
processes. In between, and this is perhaps the most
interesting case, we have the preservation of the
extensions.

\begin{description}

\item[Consequence-Preserving:] we want to obtain the
same consequences of the original theory;

\item[Extension-Preserving (Faithful):] we want a 
bijective correspondence between the extensions;

\item[Process-Preserving:] bijective correspondence between
the processes.

\end{description}

In all three cases, we assume that new variables can be
introduced, as it is common in translations between logics.
Technically, this is possible thanks to the concept of
var-equivalence \cite{lang-etal-03}.

\begin{definition}

Two formulae $\alpha$ and $\beta$ are var-equivalent \wrt\
variables $X$ if and only if $\alpha \models \gamma$ iff
$\beta \models \gamma$ for every formula $\gamma$ that only
contains variables in $X$.

\end{definition}

In plain terms, two formulae are var-equivalent if and only
if their consequences, if restricted to be formulae on a
given alphabet, are the same.

The translations we consider may introduce new variables: a
theory $\l D,W \r$ that only contains variables $X$ is
translated into a default theory $\l D',W' \r$ that contains
variables $X \cup Y$. Preservation is assumed to hold modulo
var-equivalence: preserving the extensions means that each
extension of $\l D,W \r$ is var-equivalent to an extension
of $\l D',W' \r$ \wrt\  $X$, and vice versa; preserving the
consequences means that $\l D,W \r \models \gamma$ iff $\l
D',W' \r \models \gamma$ for each formula $\gamma$ that only
contains variables in $X$. Faithful translations based on
var-equivalence of extensions have been considered by
Delgrande and Schaub~\cite{delg-scha-03} and by
Janhunen~\cite{janh-98,janh-03}.

The condition of process preservation requires a suitable
correspondence of sequences of defaults. We assume that the
defaults are numbered, both in the original and in the
generated theory. We could then enforce the consequences
being the same or being var-equivalent; this is however not
necessary as polynomial process-preserving translations will
be proved not to exist anyway.

A condition on translations that has been considered by
several authors is that of modularity
\cite{imie-87,gott-95,janh-03}, which requires only the
defaults to be translated. Formally, a default theory $\l
D,W \r$ should be translated into a theory $\l D',W' \cup W
\r$, where $\l D',W' \r$ is the result of translating $\l
D,\emptyset \r$. Another condition on translations that has
been considered in the past is that of locality
\cite{kono-88}: each sentence and each default is translated
separately. We do not consider modularity and locality in
this paper.

A requirement we impose on the translations is that of being
polynomial. There are two possible definitions of
polynomiality, depending on what is required to be
polynomial: the running time or the produced output. This
difference is important, as some translations require
exponential time but still output a polynomially large
theory. Formally, two kinds of translations are considered.

\begin{description}

\item[polynomial:] run in polynomial time;

\item[polysize:] produce a polynomially
large result.

\end{description}

The existence of translations depends both on what we want
to preserve (consequences, extensions, or processes), and
also on which computational condition we set (polynomial
time or size.) The results proved in this paper are
summarized in Table~\ref{results}.

\

\begin{table}[ht]
\begin{tabular}{l|c|c|}
& Polynomial-time & Polysize \\
\hline
Almost consequence-preserving	& yes	& yes \\
Consequence-preserving		& no	& yes \\
Faithful (extension-preserving)	& no	& yes \\
Process-preserving 		& no	& no \\
\hline
\end{tabular}
\caption{Translatability from regular to fail-safe
semantics.}
\label{results}
\end{table}

\subsection{Theories Having No Extensions}

Some semantics for default logics, like Reiter's, may not
have extensions even if $W$ is consistent. On the other
hand, all fail-safe default semantics have extensions. As a
simple result, there is no way for translating all theories
from Reiter's default logics into any fail-safe semantics in
general. This argument has been used to prove that Reiter's
default logics cannot be translated into justified or
constrained default logic by Delgrande and Schaub
\shortcite{delg-scha-03}, and that semi-normal default
theories cannot always be translated into normal default
theories by Janhunen \shortcite{janh-03}.

These results are however only consequences of the property
of having or not having extensions, not of the property of
fail-safeness. For example, we could define a default theory
to always have $[~]$ as a successful and closed process if
no other one exists, and this change would not affect the
fail-safeness of the semantics. As a result, we do not use
the possible lack of extensions to prove the impossibility
of translations.

In facts, the problem of the possible lack of extensions can
be ignored by simply assuming that the default theory to
translate has extensions. This simple assumption makes it
possible some translations that are otherwise impossible:
for example, there exists a poly-size faithful translation
from Reiter's default logics into normal or justified
default logic if we restrict to default theories having
extensions. In the rest of this paper, {\bf we only consider
theories having extensions.}

 %

\section{Translations}

In this section, we show two translations from an arbitrary
regular semantics into a normal default theory. The first
one is a poly-size faithful translation; the second one is a
polynomial translation that is ``almost''
consequence-preserving. Both translations are based on the
idea of ``simulating'' the construction of processes of the
original theory. We first show how this simulation can be
done, and then apply it for obtaining the two translations.

\subsection{Simulation of Defaults}
\label{simulate}

The reductions we show are based on simulating a regular
semantics using only normal defaults. Before going into the
technical details, we explain the basic idea of the
translation. For each default of the original theory, we
introduce two variables that represent the application of
the default in the original theory. Once the value of these
variables are set, we can use other defaults for checking
successfulness and closure of the original process. The fact
that extensions are generated only when the process is known
to be successful and closed is taken into account in two
ways:

\begin{enumerate}

\item we use new variables, and draw conclusions on the
original variables only when we know that the simulated
process is successful and closed;

\item if the simulated process is not successful or not
closed, the extension we generate is a formula $F$; the
specific choice of $F$ depends on whether we want a faithful
or an almost-consequence-preserving translation, as will be
explained in the next sections.

\end{enumerate}

Given a default theory $\l D,W \r$ and a formula $F$, both
built over the alphabet $X$, we generate a normal default
theory $\l D_u^F,W_u \r$ such that each extension of $\l D,W
\r$ is var-equivalent to an extension of $\l D_u^F,W_u \r$
\wrt\  $X$, and each extension of $\l D_u^F,W_u \r$ is
var-equivalent either to $F$ or to an extension of $\l D,W
\r$ \wrt\  $X$. Since all var-equivalences are \wrt\  $X$,
we often omit the part ``\wrt\  $X$'' in what follows.

Given $\l D,W \r$ and a specific regular semantics, the
conditions of successfulness and closure of a process $\Pi$
can be expressed as a number of consistency checks over the
formulae of $\l D,W \r$ and $\Pi$. Let
$\omega_1,\ldots,\omega_u$ be the formulae to be checked for
consistency. These formulae may depend on which defaults are
in $\Pi$, \ie, they are not exactly boolean formulae, as
they may include the propositions like $(d_i \in \Pi)$,
where $d_i \in D$. The successfulness and closure of a
process can be checked in polynomial time given the results
of these consistency checks. By a well-known result in
circuit complexity \cite{bopp-sips-90}, every polynomial
boolean function can be expressed by a circuit of polynomial
size.

The first steps of the translations are:

\begin{enumerate}

\item for each $d_i \in D$, we introduce two new variables
$c_i$ and $e_i$;

\item we introduce $m+1$ sets of new variables $X_0$, $X_1$,
\ldots, $X_m$, where $m=|D|$; each of these sets $X_i$ is in
bijective correspondence with $X$;

\item for each formula $\omega_i$, we consider the formula
$\delta_i$ that is obtained by replacing each term $(d_j \in
\Pi)$ with $c_j$ in $\omega_i$, \ie, $\delta_i =
\omega_i[(d_j \in \Pi)/c_j]$;

\item we build the circuit
$C(o_1,\ldots,o_u,c_1,\ldots,c_m)$ that encodes the
successfulness and closure tests, where the input variables
$o_1,\ldots,o_m$ represent the results of the consistency
checks and each $c_i$ encodes the presence of $d_i$ in the
process to be checked.

\end{enumerate}

We give an example of the formulae $\delta_i$ and the
circuit $C$ used for translating $\l D,W \r$ from Reiter's
semantics into normal default logic. The consistency checks
to be done are $u=2m$, where $m$ is the number of defaults.
Namely, for each default
$d_i=\frac{\alpha_i:\beta_i}{\gamma_i}$ we have to check
the consistency of the following formulae:

\begin{eqnarray*}
\omega_i &=&
W \wedge 
\left( \bigwedge_{d_j \in D} (d_j \in \Pi) \rightarrow \gamma_j \right)
\wedge \beta_i
\\
\omega_{m+i} &=&
W \wedge
\left( \bigwedge_{d_j \in D} (d_j \in \Pi) \rightarrow \gamma_j \right)
\wedge \alpha_i
\end{eqnarray*}

The formulae $\delta_i$ are obtained by replacing $d_i \in
\Pi$ with $c_i$, and are therefore the following ones:

\begin{eqnarray*}
\delta_i &=&
W \wedge
\left( \bigwedge_{d_j \in D} c_j \rightarrow \gamma_j \right)
\wedge \beta_i
\\
\delta_{m+i} &=&
W \wedge
\left( \bigwedge_{d_j \in D} c_j \rightarrow \gamma_j \right)
\wedge \alpha_i
\end{eqnarray*}

The circuit $C(o_1,\ldots,o_u,c_1,\ldots,c_m)$ encodes the
successfulness and closure of a process, provided that the
consistency of $\delta_i$ is represented by the value of
$o_i$ and the presence of $d_i$ in the process is
represented by the value of $c_i$. For Reiter's default
logic, we have:

\[
C(o_1,\ldots,o_u,c_1,\ldots,c_m) =
\bigwedge_{d_i \in D}
\Big[
(c_i \rightarrow o_i) \wedge
(\neg c_i \rightarrow (o_{m+1} \vee \neg o_i))
\Big]
\]

In words, for each default $d_i \in D$, if $d_i$ is in $\Pi$
(\ie, $c_i$ is true) then the justification of $d_i$ must be
consistent with the consequences of all defaults in $\Pi$,
(\ie, $o_i$ is true). If $d_i \not\in D$ (\ie, $c_i$ is
false), then either the precondition of $d_i$ is not
entailed (\ie, $o_{m+1}$) or its justification is not
consistent (\ie, $\neg o_i$).

What has been shown are the formulae $\delta_i$ and the
circuit $C$ for the particular case of Reiter's semantics.
Similar definitions can be given for every regular
semantics. The default theory that results from the
translation is the following one.

\[
\l D_u^F,W_u \r = \l A \cup N \cup V \cup G \cup Z, W[X/X_0] \r
\]

Let the defaults of the original theory be
$D=\{d_1,\ldots,d_m\}$, and let
$d_i=\frac{\alpha_i:\beta_i}{\gamma_i}$. The defaults of the
translated theory $D_u^F$ are defined as follows.

\begin{description}

\item[$A = \{a_i ~|~ d_i \in D\}$\rm ;] each default $a_i$
represents the application of the default $d_i$ in the
simulated process:

\[a_i =
\frac{\alpha_i[X/X_0]:\gamma_i[X/X_0] \wedge c_i \wedge e_i}
{\gamma_i[X/X_0] \wedge c_i \wedge e_i}
\]

\item[$N = \{n_i ~|~ d_i \in D\}$\rm ;] each default $n_i$
represents the choice of not applying the default $d_i$:

\[
n_i =
\frac{:\neg c_i \wedge e_i}{\neg c_i \wedge e_i}
\]

\item[$V=\{v_1,\ldots,v_u\}$\rm ;] each $v_i$ relates the
consistency of $\delta_i$ with the value of the variable
$o_i$:

\[
v_i = \frac{e_1 \wedge \cdots \wedge e_m:
\delta_i[X/X_i] \wedge o_i \wedge t_i}
{\delta_i[X/X_i] \wedge o_i \wedge t_i}
\]

\item[$G=\{g_1,\ldots,g_u\}$\rm ;] each $g_i$ relates the
inconsistency of $\delta_i$ with the value of $o_i$:

\[
g_i = \frac{e_1 \wedge \cdots \wedge e_m \wedge \neg \delta_i[X/X_i]:
\neg o_i \wedge t_i}
{\neg o_i \wedge t_i}
\]

\item[$Z=\{z_1,z_2\}$\rm ;] these two defaults are used to
compute the result of the circuit $C$ and to ``output'' the
generated extension or $F$ accordingly:

\begin{eqnarray*}
z_1 &=& 
\frac{e_1 \wedge \cdots \wedge e_m \wedge 
t_1 \wedge \cdots \wedge t_u \wedge
C(o_1,\ldots,o_u,c_1,\ldots,c_m):
W \wedge \bigwedge_{d_i \in D} (c_i \rightarrow \gamma_i)}
{W \wedge \bigwedge_{d_i \in D} (c_i \rightarrow \gamma_i)}
\\
z_2 &=&
\frac{e_1 \wedge \cdots \wedge e_m \wedge
t_1 \wedge \cdots \wedge t_u \wedge 
\neg C(o_1,\ldots,o_u,c_1,\ldots,c_m):F}
{F}
\end{eqnarray*}

\end{description}

We use the following abbreviations: $E$ for $e_1 \wedge
\cdots \wedge e_m$ and $T$ for $t_1 \wedge \cdots \wedge
t_u$. The preconditions of the defaults in $D_u^F$ have been
defined so that the defaults of $A \cup N$ have to be
applied first; then, the defaults of $V \cup G$ can be
applied; the defaults of $Z$ can be applied only at the end.
We prove a number of lemmas relating the processes of the
theory $\l D_u^F,W_u \r$ with the processes of $\l D,W \r$.

\begin{lemma}
\label{first}

If $\Pi$ is a successful and closed process of $\l D_u^F,
W_u \r$, then $\Pi$ contains exactly one among $a_i$ or
$n_i$ for each $i$, and contains them only in the first $m$
positions.

\end{lemma}

\proof Since all defaults in $D_u^F \backslash (A \cup N)$
contain $E$ as a precondition, and $E$ is only made true
once $m$ defaults of $A \cup N$ are applied, the first $m$
defaults of $\Pi$ are in $A \cup N$. If both $a_i$ and $n_i$
are in $\Pi$, then $\Pi$ is not successful, as the
consequences of these defaults contradicts each other. If
neither $a_i$ nor $n_i$ are in $\Pi$, then the default $n_i$
is applicable; therefore, $\Pi$ is not a closed process.
Finally, since exactly one between $a_i$ and $n_i$ is in the
first $m$ positions of $\Pi$, no other defaults of $A \cup
N$ can be in a position of $\Pi$ after the $m$-th.~\qed

In words, the processes of $\l D_u^F,W_u \r$ begin with the
application of exactly one between $a_i$ or $n_i$ for each
$i$. After that, no other default of $A \cup N$ can be
applied. The idea is that the truth value of $c_i$ reflects
the application of $d_i$ in the default theory $\l D,W \r$.
Given a process $\Pi$ of $\l D_u^F,W_u \r$, we define the
``simulated process'' $O(\Pi)$ to be the following process
of $\l D,W \r$:

\[
O(\Pi) = [d_{i_1},\ldots,d_{i_k}]
\mbox{ where $a_{i_1},\ldots,a_{i_k}$ is the
sequence of defaults $a_i$ of $\Pi$}
\]

We prove that $O(\Pi)$ is a process of the original theory
$\l D,W \r$.

\begin{lemma}
\label{process-process}

For every process $\Pi$ of $\l D_u^F, W_u \r$, $O(\Pi)$ is a
process of $\l D,W \r$.

\end{lemma}

\proof By assumption, $\Pi$ is a process of $\l D_u^F, W_u
\r$. As a result, $W_u \cup \cons(\Pi)$ is consistent and
$W_u \cup \cons(\Pi[d]) \models \prec(d)$ for every $d \in
\Pi$. These two conditions imply the same ones on $W$ and
$O(\Pi)$ once the inverse substitution $[X_0/X]$ has been
applied.~\qed

The converse of this lemma also holds.

\begin{lemma}
\label{process-successful}

For each process $\Pi'$ of $\l D,W \r$, there exists a
successful and closed process $\Pi$ of $\l D_u^F, W_u \r$
such that $O(\Pi)=\Pi'$.

\end{lemma}

\proof Let $\Pi_1$ be the process obtained by replacing each
$d_i$ with $a_i$ in $\Pi'$ and adding all $n_i$'s such that
$d_i \not\in \Pi'$. This process $\Pi_1$ is successful, and
$O(\Pi_1)=\Pi'$. Since $\l D_u^F, W_u \r$ is normal, $\Pi_1$
is the prefix of a successful and closed process $\Pi_1
\cdot \Pi_2$. Since $\Pi_1$ already contains $m$ defaults,
no default of $\Pi_2$ is in $A \cup N$. As a result,
$O(\Pi_1 \cdot \Pi_2)$ is equal to $O(\Pi_1)$, which is in
turn equal to $\Pi'$.~\qed

Together, these lemmas prove that the processes of $\l D,W
\r$ are in correspondence with the successful and closed
processes of $\l D_u^F,W_u \r$. Namely, each process $\Pi'$
of $\l D,W \r$ corresponds to some successful and closed
processes $\Pi$ of $\l D_u^F,W_u \r$ such that
$O(\Pi)=\Pi'$, and for each successful and closed processes
$\Pi$ of $\l D_u^F,W_u \r$, it holds that $O(\Pi)$ is a
process of $\l D,W \r$. What is still missing is the effect
of the successfulness and closure of the process of $\l D,W
\r$ on the corresponding processes of $\l D_u^F,W_u \r$. We
prove two preliminary lemmas

\begin{lemma}
\label{second}

Every successful and closed process $\Pi$ of $\l D_u^F,W_u
\r$ contains either $v_i$ or $g_i$, but not both, for every
$i$, in the positions of $\Pi$ from the $m+1$-th to the
$m+u$-th.

\end{lemma}

\proof As for Lemma~\ref{first}, after applying the first
$m$ defaults the formula $E$ is true but $T$ is not. As a
result, we can only apply defaults in $V \cup G$. The rest
of the proof is like that of Lemma~\ref{first}.~\qed

\begin{lemma}
\label{checks}

If $\Pi'$ is a process of $\l D,W \r$ and $\Pi$ is a
successful and closed process of $\l D_u^F,W_u \r$ such that
$O(\Pi)=\Pi'$, then $\cons(\Pi)$ entails $o_i$ or $\neg o_i$
depending on whether $\Pi'$ satisfies $\omega_i$.

\end{lemma}

\proof By Lemma~\ref{first} and Lemma~\ref{second}, the
first $m$ defaults of $\Pi$ are in $A \cup N$ and the next
$u$ defaults are in $V \cup G$. Since $O(\Pi)=\Pi'$, $a_j
\in \Pi$ if and only if $d_j \in \Pi'$, and $n_j \in \Pi$ if
and only if $d_j \not\in \Pi'$. As a result, $\cons(\Pi)$
contains either $c_j$ or $\neg c_j$, depending on whether
$d_j \in \Pi'$. Since the value of each $c_j$ indicates the
presence of $d_j \in \Pi'$, the satisfiability of the
formulae $\delta_i$ corresponds to the satisfaction of the
conditions $\omega_i$.

By construction, the default $v_i$ is only applicable if
$\delta_i$ is consistent, which means that $\Pi'$ satisfies
the condition $\omega_i$. For the same reason, $g_i$ is
applicable only if $\omega_i$ is not satisfied. As a result,
$\cons(\Pi)$ contains either $o_i$ or $\neg o_i$ depending
on whether $\Pi'$ satisfies the condition $\omega_i$.~\qed

We now establish a correspondence on the conditions of
successfulness and closure.

\begin{lemma}
\label{closed}

If $\Pi'$ is a successful and closed process of $\l D,W \r$
and $\Pi$ is a successful and closed process of $\l D_u^F,
W_u \r$ such that $O(\Pi)=\Pi'$, then $W_u \wedge
\cons(\Pi)$ and $W \wedge \cons(\Pi')$ are var-equivalent
\wrt\ $X$.

\end{lemma}

\proof The precondition of $z_1$ and $z_2$ include $E \wedge
T$ and either $C(\ldots)$ or $\neg C(\ldots)$, respectively.
By Lemma~\ref{first} and Lemma~\ref{second}, $E \wedge T$ is
entailed by $\cons(\Pi)$. By Lemma~\ref{checks}, the truth
value of each $o_i$ is related to the process $\Pi'$
satisfying the condition $\omega_i$; since $\Pi'$ is
successful and closed, $C$ evaluates to true. Therefore,
$z_1$ is applicable while $z_2$ is not. Since $\Pi$ is
successful and closed, $z_1$ is in $\Pi$. The
var-equivalence of $W_u \wedge \cons(\Pi)$ with $W \wedge
\cons(\Pi')$ is due to the fact that the consequence of
$z_1$ is equivalent the latter formula after having
replaced each $c_i$ with either true or false, depending on
whether $a_i \in \Pi$.~\qed

The converse of this lemma also holds.

\begin{lemma}
\label{nonclosed}

If $\Pi'$ is a process of $\l D,W \r$ that is either not
successful or not closed and $\Pi$ is a successful and
closed process of $\l D_u^F, W_u \r$ such that
$O(\Pi)=\Pi'$, then $W_u \wedge \cons(\Pi)$ is
var-equivalent to $F$ \wrt\  $X$.

\end{lemma}

\proof Same as the proof of the previous theorem, but $C$
this time evaluates to false. Therefore $\Pi$ includes
$z_2$, which has $F$ as a consequence.~\qed

These lemmas establish a correspondence between the
extensions of the original and generated theories.

\begin{theorem}

Every extension of $\l D,W \r$ is var-equivalent to an
extension of $\l D_u^F,W_u \r$ and every extension of $\l
D_u^F,W_u \r$ is var-equivalent either to $F$ or to an
extension of $\l D,W \r$.

\end{theorem}

\proof If $\Pi'$ is a successful and closed process of $\l
D,W \r$, then there exists a closed and successful process
$\Pi$ of $\l D_u^F,W_u \r$ such that $O(\Pi)=\Pi'$ by
Lemma~\ref{process-successful}. By Lemma~\ref{closed}, it
holds that the extension generated by $\Pi$ is
var-equivalent to that generated by $\Pi'$.

If $\Pi$ is a closed and successful process of $\l
D_u^F,W_u \r$, by Lemma~\ref{process-process} $O(\Pi)$ is a
process of $\l D,W \r$. If $O(\Pi)$ is successful and
closed, by Lemma~\ref{closed}, the extension generated by
$\Pi$ is var-equivalent to that generated by $\Pi'$. If
$O(\Pi)$ is either not successful or not closed, by
Lemma~\ref{nonclosed}, the extension generated by $\Pi$ is
var-equivalent to $F$.~\qed

 %

\subsection{Faithful (Extension-Preserving) Translations}

Delgrande and Schaub \cite{delg-scha-03} have proved that
Reiter's default logic cannot be translated into justified
default logic. Janhunen \cite{janh-03} has proved that
semi-normal defaults cannot be translated into normal
defaults under Reiter's semantics. Both results imply the
impossibility of translating Reiter's semantics into a
fail-safe semantics. However, both proofs are based on the
possible lack of extensions in Reiter's semantics. We
therefore investigate whether theories having extensions
under Reiter's semantics can be faithfully translated into a
fail-safe default theory, namely, normal default logic.

A faithful, but exponential, translation from every regular
default logic semantics into normal default logics always
exists: if the extensions of the original theories are
obtained by the deductive closure of the formulae in
$\{E_1,\ldots,E_m\}$, the following theory is a faithful
translation of it into normal default logic, where the
$e_i$'s are new variables.

\[
T = 
\ll
\left\{
\left.
\frac{:e_i \wedge E_i \wedge \bigwedge_{j=1,\ldots,m ~ j \not= i} \neg e_j}%
{e_i \wedge E_i \wedge \bigwedge_{j=1,\ldots,m ~ j \not= i} \neg e_j}
\right|
i=1,\ldots,m 
\right\},
\emptyset
\rr
\]

This theory contains exactly one default for each extension
of the original theory. Since these defaults are normal, and
their justifications are inconsistent with each other, the
successful and closed processes of this theory are exactly
the sequences composed of a single default. The generated
extensions are exactly the same (modulo var-equivalence) of
the original theory.

The problem with this translation is not only that it is
exponential: even worst, once the set of all extensions
$\{E_i\}$ has been determined, there is no reason for using
default logic, as we can simply use propositional logic
instead.

Polynomial translations can be of two kinds: either the new
theory can be built in polynomial time, or it has polynomial
size. The first condition implies the second, but not vice
versa. A translation from a regular semantics into normal
default logic exists if we only require the result of the
translation to be polynomial in size.

\begin{theorem}

For every regular semantics there exists a poly-size
faithful (extension-preserving) translation that maps all
default theories that have extensions into normal default
theories.

\end{theorem}

\proof The simulation shown in Section~\ref{simulate} is
``almost'' an extension-preserving translation. Indeed, all
extensions of the original theory are translated into
extensions of the generated theory. On the other hand,
processes of the original theory that do not generate
extensions correspond to processes that generate $F$ as an
extension.

We can build a faithful translation as follows: first, we
determine a single successful and closed process $\Pi$ of
the original theory. We then translate the default theory
$\l D,W \r$ into the simulating theory $\l D_u^F, W_u \r$ in
which $F=W \wedge \cons(\Pi)$. This is a faithful
translation because each successful and closed process of
the original theory $\l D,W \r$ corresponds to a successful
and closed process of the theory $\l D_u^F,W_u \r$ generating
the same extension (modulo var-equivalence). The processes
of $\l D,W \r$ that are either not successful or not closed
correspond to processes of $\l D_u^F,W_u \r$ that generate an
extension that is var-equivalent to $F$. Since $F$ is an
extension of the original theory, this translation is
faithful.~\qed

The translation of this theorem is not polynomial-time, as
it requires the generation of at least one successful and
closed process of the original theory. On the other hand,
such a process has always polynomial size. The result of the
translation is therefore always of polynomial size. As will
be shown later in the paper, no consequence-preserving
polynomial-time translation exists. This implies that no
faithful and polynomial-time translation exists as well.

 %

\subsection{Almost-Consequence-Preserving Translations}

A simple computational argument shows that any default
theory can be translated into a normal default theory in
polynomial time if we admit the queries to be translated as
well, \ie, $\l D,W \r \models q$ in a regular semantics if
and only if $\l D',W' \r \models q'$, where $D'$ is normal.
Indeed, query answering is in \P{2}\  for all regular
default logics and \P{2}-hard for Reiter's default logic
even in the restriction to normal theories
\cite{gott-92-b,stil-92-b}. However, for a translation to be
``exactly'' consequence-preserving $q'$ should be the same
as $q$.

A consequence-preserving translation is easy to give if we
allow an exponential blow up of the theory: if the
extensions of $\l D,W \r$ are $\{E_1,\ldots,E_m\}$, the
skeptical consequences of $\l D,W \r$ are exactly the
classical consequences of $E_1 \vee \cdots \vee E_m$, which
are also the skeptical consequences of the default theory
$\l \emptyset, \{E_1 \vee \cdots \vee E_m\} \r$. Ben-Eliyahu
and Dechter \cite{bene-dech-96} defined a better translation
from default logic into propositional logic, which can be
polynomial even if the number of extensions is exponential.
Translations from default logic into propositional logic are
however known to be exponential in the worst case due to the
different complexity of the semantics. We now concentrate on
polynomial translations.

The case of default theories having no extensions has
already been considered, so we restrict to theories that
have extensions. We have already shown a faithful
translation that is poly-size: this translation clearly
preserves the consequences as well. In this section, we show
a translation from every regular semantics into normal
default logic that is:

\begin{enumerate}

\item polynomial-time;

\item ``almost'' consequence preserving: if $\l D, W \r$ is
the original theory and $\l D',W' \r$ is the result of the
translation, then $\l D, W \r \models q$ if and only if $\l
D',W' \r \models a \vee q$, where $a$ is a new variable
created by the translation.

\end{enumerate}

The translation is based on the theory that simulates the
process construction of the original theory. As we have
already noticed, the only problem with this simulation is
that the processes of the original theory that either are
not successful or not closed correspond to successful and
closed processes in the simulating theory. Since these
processes generate $Cn(F)$ as an extension, all we have to
do is to specify a value of $F$ that do not affect
entailment.

The trick we use is to translate $q$ into $a \vee q$ and to
set $F=a$. If we use the skeptical semantics, the extensions
of the simulating theory that do not correspond to
extensions of the original theory imply $a$, which in turn
implies $a \vee q$; as a result, they do not affect the
consequences of the theory.

\begin{theorem}

For every regular default logic there exists a
polynomial-time translation that maps a default theory $\l
D,W \r$ that has extensions into a normal default theory $\l
D',W' \r$ such that $\l D,W \r \models q$ if and only if $\l
D',W' \r \models a \vee q$, where $a$ is a new variable.

\end{theorem}

The formula $F$ and the way in which queries are translated
are chosen in such a way the extensions of $\l D',W' \r$
that do not correspond to extensions of the original theory
$\l D,W \r$ are irrelevant to the specific query evaluation
mechanism. As a result, if we are interested into credulous
entailment, we can use $F=\neg a$ and translate a query $q$
into $a \wedge q$

The question of whether the addition of $a$ to the queries
is necessary depends on the kind of translation used: we
have already shown a faithful (and, therefore,
consequence-preserving) translation that is poly-size. We
will show that no polynomial-time consequence-preserving
(\ie, that do not modify queries at all) translation exists
unless part of the polynomial hierarchy collapses.

 %

 %

\section{Impossibility of Translations}

In this section, we show that some translations are
impossible: namely, there is no polynomial-time exact
consequence-preserving translation and no polynomial-time or
polysize process-preserving translation from Reiter's
default logics into any fail-safe default logic.

\subsection{Consequence-Preserving Translations}

We have already shown a poly-size faithful translation and a
polynomial-time almost-consequence-preserving translation.
We prove that no polynomial-time reduction that preserves
the consequences exactly exists. To this end, we show a
problem that is hard for Reiter's default logic but easy for
all fail-safe default semantics. We cannot use a problem
that has already been analyzed in the past (such as
entailment or model checking) because these problems have
the same complexity for Reiter's and for some fail-safe
semantics.

For all fail-safe semantics, generating an extension is
relatively easy, as it can be done by applying defaults
until the process is closed. This property can be used to
define a problem that is hard for Reiter's semantics but
easy for all fail-safe ones: if it is known that either all
extensions imply $a$ or all extensions imply $\neg a$, then
a single arbitrary extension suffices to check whether $a$
is entailed. In turns, the assumption that all extensions
imply $a$ or all extensions imply $\neg a$ is equivalent to
the assumption that the default theory implies either $a$ or
$\neg a$. We prove that entailment is hard for Reiter's
default logic even under this assumption.

\begin{theorem}

The problem of checking whether $\l D,W \r \models a$ in
Reiter's default logic is $\S{2} \cap \P{2}$-hard even if
$\l D,W \r$ implies either $a$ or $\neg a$, and it has
extensions.

\end{theorem}

\proof Let $P$ be a problem in $\S{2} \cap \P{2}$. We
reduce the problem of telling whether $x \in P$ to
the problem $T \models a$, where $T$ is a default
theory that either implies $a$ or it implies $\neg a$.

Since $P$ it is in \S{2}, the question $x \in P$ can be
reduced to the problem of checking the existence of
extensions of a default theory with an empty background
theory $\l D_p, \emptyset \r$ \cite{gott-92-b}. Since $P$ is
in \P{2}, its complementary problem is in \S{2}\  as well.
As a result, the question $x \not\in P$ can therefore be
reduced to the existence of extensions of another theory $\l
D_n, \emptyset \r$. Therefore, $x \in P$ if and only if $\l
D_p, \emptyset \r$ has extensions while $\l D_n, \emptyset
\r$ has not, and vice versa if $x \not\in P$.

Let $a$ be a variable that is mentioned neither in $D_p$
nor in $D_n$. The default theory we use is the following one:

\begin{eqnarray*}
T &=& \l D,\emptyset \r
\\
& \mbox{\hbox to 0pt{where}}
\\
D &=&
\left\{
\frac{:a}{a} ,~
\frac{:\neg a}{\neg a}
\right\}
\cup
\left\{ \left.
\frac{a \wedge \prec(d):just(d)}{\cons(d)}
\right| d \in D_p
\right\}
\cup
\left\{ \left.
\frac{\neg a \wedge \prec(d):just(d)}{\cons(d)}
\right| d \in D_n
\right\}
\end{eqnarray*}

The only two defaults that can be applied from the
background theory are the first two. They cannot be applied
together, however. Once the first one is applied, the theory
becomes equivalent to $\l D_p, \{a\} \r$, while the
application of the second one makes it equivalent to $\l
D_n, \{\neg a\} \r$. Since the existence of extensions for
these two theories are related to the question $x \in P$, we
have:

\begin{enumerate}

\item if $x \in P$, all extensions of $T$ imply $a$;

\item if $x \not\in P$, all extensions of $T$ imply $\neg a$.
 
\end{enumerate}

As a result, either $T \models a$ or $T \models \neg a$; in
particular, $T \models a$ if and only if $x \in P$.~\qed

The same problem is relatively easy for every fail-safe
default semantics. Indeed, to solve it we only need to
generate an extension and to check whether it implies $a$ or
$\neg a$: since all extensions are the same as for the
entailment of $a$ and $\neg a$, checking one extension
suffices. Generating an arbitrary extension can be done
easily in every fail-safe semantics.

\begin{theorem}

Checking whether $T \models a$ is in \D{2}\  for every
fail-safe semantics, if either $T \models a$ or $T \models
\neg a$.

\end{theorem}

\proof Since either all extensions of $T$ imply $a$ or all
extensions of $T$ imply $\neg a$, we can check whether $T
\models a$ by finding a single extension $E$ of $T$ and then
checking whether $E \models a$. Finding one extension $E$ is
easy because the semantics is fail-safe. If $\Pi$ is a
successful process that is not closed, then there exists
$\Pi'$ such that $\Pi \cdot \Pi'$ is successful and closed.
By the antimonotonicity of successfulness, if $d$ is the
first default of $\Pi'$, \ie, $\Pi'=[d] \cdot \Pi''$, then
$\Pi \cdot [d]$ is successful. As a result, if $\Pi$ is
successful then it is either closed or there exists a
default $d$ such that $\Pi \cdot [d]$ is successful. We can
therefore start with $\Pi = [~]$, which is successful. At
each step, if $\Pi$ is closed, we can check whether
$\cons(\Pi) \models a$. Otherwise, there exists $d$ such
that $\Pi \cdot [d]$ is successful. We set $\Pi=\Pi \cdot
[d]$, and continue. This algorithm must necessarily end up
with a successful and closed process.

This algorithm only takes a polynomial number of steps if we
have access to an \np-oracle. Indeed, all we have to do is
to check closure of $\Pi$ and successfulness of $\Pi \cdot
[d]$ at each step; these conditions can be verified in
polynomial time by letting the \np-oracle perform the
consistency tests.~\qed

As a result of these two theorems, no polynomial-time
consequence-preserving translation exists from Reiter's
semantics to an arbitrary fail-safe semantics, unless $\S{2}
\cap \P{2} = \D{2}$. The following theorem shows that even a
polynomial number of calls to an \np-oracle do not suffice
to translate from Reiter's semantics to any fail-safe one.

\begin{theorem}

If there exists an (exact) consequence-preserving
translation from Reiter's semantics into any fail-safe
semantics that only requires a polynomial number of calls to
an \np-oracle, then $\S{2} \cap \P{2} = \D{2}$.

\end{theorem}

\proof It such a translation exists, then for any $P \in
\S{2} \cap \P{2}$ we could translate the question $x \in P$
into the question $T \models a$ under Reiter's semantics
where either $T \models a$ or $T \models \neg a$. In turns,
with a polynomial number of calls to the oracle we can
translate the question into the same question for a
fail-safe semantics, where it can be solved with a
polynomial number of other calls to the oracle.~\qed

Since no translation employing a polynomial number of calls
to an \np-oracle exists, no polynomial-time translation
exists either. Since the theorem has been proved using only
theories in which all extensions have the same behavior
\wrt\  the query (either they all entail it, or they all
entail their negation,) this result holds for both skeptical
and credulous reasoning.

The impossibility of polynomial-time faithful translations
is a consequence of the above theorem: a faithful
translation is also a consequence-preserving translation,
and cannot therefore be polynomial-time.

\begin{theorem}

If there exists a faithful translation from Reiter's default
logic into any fail-safe default logic that only requires a
polynomial number of calls to an \np-oracle, then $\S{2}
\cap \P{2} = \D{2}$.

\end{theorem}

 %

\subsection{Process-Preserving Translations}

A process-preserving translation is a translation that not
only preserves the extensions, but also the processes of a
default theory. Clearly, we cannot enforce the processes to
be exactly the same, otherwise the two theories would have
the same defaults. Therefore, we only impose that there is a
one-to-one correspondence between the defaults of the
original and generated theories, such that the processes of
the original theory matches the processes of the generated
theory thanks to this correspondence.

An easy way for creating this correspondence is to assume
that the first part $D$ of a default theory $\l D,W \r$
is a sequence of defaults rather than a set. In other words,
we add an enumeration on the defaults so that we can write
$D=\{d_1,\ldots,d_m\}$. A process-preserving translation is
a function that maps a default theory $\l
\{d_1,\ldots,d_m\},W \r$ into another default theory $\l
\{d_1',\ldots,d_m'\},W' \r$ with the same number of defaults,
and such that $[d_{i_1},\ldots,d_{i_r}]$ is a successful and
closed process of the first theory if and only if
$[d_{i_1}',\ldots,d_{i_r}']$ is a successful and closed
process of the second one.

We prove that there is no polynomial-time or polysize
process-preserving translation from Reiter's default logics
into any fail-safe default logic. To this aim, we show a
problem that is hard in Reiter's default logic but easy in
all fail-safe semantics.

\begin{definition}[Completability of Process]

Given a default theory $\l D,W \r$ and a sequence of
defaults $\Pi$, check whether there exists a successful a
complete process $\Pi \cdot \Pi'$.

\end{definition}

The following theorem characterizes the complexity of
completability of processes for Reiter's semantics.

\begin{theorem}

The problem of completability of processes is \S{2}-complete
in Reiter's default logic even for theories that have
extensions.

\end{theorem}

\proof Membership: guess a sequence of defaults $\Pi'$ and
check whether $\Pi \cdot \Pi'$ is a successful and closed
process.

Hardness: we reduce the problem of existence of extensions
to this one. Given a theory $\l D,W \r$ we build
the theory $\l D',W \r$, where $D'=\{d_n,d_p\} \cup D''$
and $d_n$, $d_p$, and $D''$ are defined as follows.

\begin{eqnarray*}
d_n &=& \frac{:\neg a}{\neg a} \\
d_p &=& \frac{:a}{a} \\
D'' &=&
\left\{ \left.
\frac{a \wedge \prec(d):\just(d)}{\cons(d)}
\right| d \in D \right\}
\end{eqnarray*}

This theory, as required, has one extension: the one
generated by the process $[d_n]$. The other processes, if
any, are made of $d_p$ followed by the defaults that
corresponds to the successful and closed processes of $\l
D,W \r$. As a result, the consistent process $[d_p]$ can be
extended to form a consistent and closed process if and only
if $\l D,W \r$ has extensions.~\qed

The problem of completability of extensions is relatively
easy for all fail-safe semantics, as it amounts to checking
whether $\Pi$ is a successful process. By definition,
indeed, any successful process is either closed or can be
extended to form a successful process. Moreover, if $\Pi$ is
not a process, or it is not successful, then it cannot be
extended to generate a successful process thanks to the
anti-monotonicity of successfulness.

As a result, completability of processes is equivalent to
verifying whether a sequence of defaults is a successful
process, which is a problem in \D{2}. For the fail-safe
semantics defined in the literature, the problem is even
simpler, as it is in \Dp. We prove that is hard for the same
class for justified default logic for the sake of
completeness.

\begin{theorem}

Checking whether $\Pi$ is a successful process is in \D{2}\
for every regular default semantics, and is \Dp-complete for
justified default logic.

\end{theorem}

\proof The conditions of $\Pi$ being a process and being
successful can be computed in polynomial time once a number
of consistency tests have been performed. The problem is
therefore in \D{2}. For the case of justified default logic,
these consistency tests are independent to each other, that
is, the formulae to check do not depend to the results of
the other tests. As a result, the problem is in \Dp.

The hardness result is an obvious consequence of the fact
that the applicability of a single default is hard: the
problem \sat-\unsat, \ie, checking whether a pair of
formulae $\l \alpha, \beta \r$ is composed of a satisfiable
formula $\alpha$ and an unsatisfiable formula $\beta$, is
\Dp-hard. This problem can indeed be reduced to the problem
of checking whether $[d]$ is a successful and closed process
of the theory below:

\[
\ll \left\{ \frac{\neg \beta:\alpha}{\alpha} \right\} ,~
\emptyset \rr
\]

The sequence of defaults $[d]$ is indeed a successful
process if and only if $\neg \beta$ is valid (that is,
$\beta$ is inconsistent) and $\alpha$ is consistent. As a
result, the problem is \Dp-hard.~\qed

Suppose that there exists a process-preserving translation
from Reiter's default logic to any fail-safe default logics.
We can then solve the problem of completability of a process
$\Pi$ in Reiter's default logic by simply translating both
the theory and the process, and then solving the problem in
the fail-safe default semantics. This would imply that
\S{2}=\D{2}.

This result can be strengthened to polysize translations. We
can indeed prove that the problem of completability of
processes does not simplify thanks to a preprocessing phase.
This is proved by showing that the problem of completability
of processes is \nucS{2}-hard for Reiter's default logics,
and cannot therefore be ``compiled to'' \D{2}. The class
\nucS{2}\  has been introduced by Cadoli et al.
\cite{cado-etal-02,libe-01} to characterize the complexity
of problems when preprocessing of problem is allowed. We
omit the details here, and refer the reader to the papers by
Cadoli et al. \cite{cado-etal-02,libe-01}.

\begin{theorem}

The problem of completability of processes is
\nucS{2}-complete in Reiter's default logic, where the
default theory is the fixed part of the instance.

\end{theorem}

\proof We adapt the reduction by Gottlob, as follows: given
a formula $\exists X \forall Y ~.~ \neg \phi$, where
$|X|=|Y|=n$ and $\phi$ contains only clauses of three
literals, let $A=\{\gamma_1,\ldots,\gamma_m\}$ be the set of
all clauses of three literals over the alphabet $X \cup Y$;
we build a default theory and a successful process of it as
follows:

\begin{eqnarray*}
T &=&
\ll
\{p_i, n_i ~|~ 1 \leq i \leq m \}
\cup
\{a_i, b_i ~|~ 1 \leq i \leq n \}
\cup
\{f\},
W
\rr
\\
&& where: \\
&& p_i = \frac{:c_i \wedge e_i}{c_i \wedge e_i} \\
&& n_i = \frac{:\neg c_i \wedge e_i}{\neg c_i \wedge e_i} \\
&& a_i = \frac{:x_i \wedge f_i}{x_i \wedge f_i} \\
&& b_i = \frac{:\neg x_i \wedge f_i}{\neg x_i \wedge f_i} \\
&& f = \frac{\bigwedge e_i \wedge \bigwedge f_i:
\{c_i \rightarrow \gamma_i ~|~ \gamma_i \in A\}}{\false} \\
&& W=\emptyset
\\
\Pi &=&
[d_1,\ldots,d_m]
\\
&& \mbox{where $d_i$ is $p_i$ if $\gamma \in \phi$ and $n_i$
otherwise}
\end{eqnarray*}

The sequence $\Pi$ is always a successful process. Moreover,
$W \wedge \cons(\Pi)$ implies all variables $e_i$ and either
$c_i$ or $\neg c_i$. Namely, $c_i$ is entailed if $p_i$ is
in $\Pi$, and $\neg c_i$ is entailed otherwise. We can
therefore replace each $c_i$ such that $\gamma_i \in \phi$
with $\true$ and each $c_i$ such that $\gamma_i \not\in
\phi$ with $\false$. The default $f$ therefore simplifies
to:

\[
f' = \frac{\bigwedge f_i:
\{\gamma_i ~|~ \gamma_i \in \phi\}}{\false}
\]

As shown by Gottlob, the resulting theory has extensions if
and only if $\exists X \forall Y ~.~ \neg \phi$. This is
therefore a polynomial translation from $\exists\forall$QBF
into the problem of completability of a process. Moreover,
the default theory only depends on the number of variables
of the QBF, while the process is the only part that depends
on the specific $\phi$. As a result, this is a
\nuc-reduction, and proves that the problem of
completability of processes is \nucS{2}-hard. Since the
problem is in \S{2}, it is in \nucS{2} as well. As a result,
the problem is \nucS{2}-complete.~\qed

This theorem proves that no polysize process-preserving
translation from Reiter's semantics into any fail-safe
semantics exists. Indeed, if this were the case, we could
translate any theory from Reiter's default logics into a
theory that has the same successful and closed processes
under a fail-safe semantics. This would prove that the
problem is in \nucD{2}, which implies that
$\nucP{2}=\nucS{2}$, and therefore the polynomial hierarchy
collapses \cite{cado-etal-02}.

 %

 %

\section{Conclusions}

The possibility for a default logic semantics to make a
sequence of applicable defaults to fail can be seen as a
semantical drawback or as a computational feature. In this
paper, we have studied how much a semantics gains from this
ability to generate failures. In particular, we have
restricted to the case of theories having extensions, and
have shown that translations from fail-prone to fail-safe
semantics are possible or not depending on the constraints
that are imposed on the translation. In particular, a
translation that preserves the extensions or the skeptical
consequences and produces a polynomial sized result exists,
while a polynomial-time translation does not. We have also
considered more liberal (almost-consequence-preserving) and
more restrictive (process-preserving) constraints on the
translations.

The main results of this paper imply that the ability of
failing can only give an advantage in terms of translation
time (\eg, not all Reiter's theories can be faithfully
translated into normal or justified theories in polynomial
time), but not in terms of expressibility (\eg, for every
Reiter's theory there exists an equivalent normal or
justified theory of polynomial size.) This distinction is
important, because it shows that fail-prone semantics are
better than fail-safe ones in solving problems by
translating them into default logic, but are not in terms of
which domains can be encoded in polynomial space. In short,
the possible failure of processes is a computational
advantage, but not an expressiveness advantage.

These results hold only under some assumptions: the
reductions are constrained to be polynomial (either in time
or space) but can introduce new variables. Moreover, we only
consider theories that have extensions, and prove the
existence of translations only for normal default logic,
which we considered a prototypical fail-safe semantics.
These assumptions make the results of this paper
incomparable to what proved in two similar works:

\begin{enumerate}

\item Delgrande and Schaub \cite{delg-scha-03} have shown
reductions from other variants of default logics into
Reiter's; they take into account theories having no
extensions, and limit to the specific case of justified,
constrained, and rational default logic;

\item Janhunen \cite{janh-03} has shown that Reiter's
default logic can be translated into semi-normal default
logic; translations are assumed not only faithful but also
modular, and theories having no extensions are taken into
account.

\end{enumerate}

Both these two works consider polynomial faithful
translations with new variables. The results presented in
our paper are more general than the ones above in the sense
that we proved the existence of translations from an
arbitrary regular default theory into a fail-safe one and
the non-existence of a translation from Reiter's semantics
into an arbitrary fail-safe one.

Some apparent contradictions between the results proved in
this paper and those by Delgrande and Schaub and by Janhunen
are due to the fact that we only consider default theories
having extensions. This is why, for example, some results
about the impossibility of translations by Delgrande and
Schaub \cite[Theorem~6]{delg-scha-03} and by Janhunen
\cite[Theorem~5]{janh-03}, which relies on the possible
non-existence of extensions in Reiter's semantics, are not
in contradiction with our results on the existence of such
reductions. These apparent contradictions show that some
translations are only impossible because of the possible
lack of extensions, and become possible as soon as theories
having no extensions are excluded from consideration.

An interesting question left open by the present work is
whether the comparison between fail-safe and fail-prone
semantics can be extended to logics that are not based on
defaults. Clearly, a suitable definition of failure is
needed; however, it seems somehow natural to consider
propositional circumscription \cite{lifs-94} as a fail-safe
non-monotonic logic (we add negative literals to a theory as
far as possible, but never retract an added literal) and
autoepistemic logic \cite{moor-85} as a fail-prone one (we
can ``generate'' a conclusion $x$ by means of a formula like
$\Box x \rightarrow x$ but then retract the conclusion, if a
condition $F$ is met, by means of a formula like $F
\rightarrow \neg x$.)

Finally, we note that frameworks for comparing propositional
knowledge representation formalisms have been given by
Cadoli et al. \shortcite{cado-etal-00} and by Penna
\shortcite{penn-00}. The translations considered in these
framework are allowed to translate queries (or models),
while the only translation of queries admitted in this paper
is the addition of a literal to queries, \ie, $q$ is
translated into~$a \vee q$.

 %

\let\sectionnewpage=\relax
\bibliographystyle{alpha}

\begin{thebibliography}{CMMT95}

\bibitem[Ant99]{anto-99}
G.~Antoniou.
\newblock A tutorial on default logics.
\newblock {\em {ACM} Computing Surveys}, 31(4):337--359, 1999.

\bibitem[AS94]{anto-sper-94}
G.~Antoniou and V.~Sperschneider.
\newblock Operational concepts of nonmonotonic logics, part 1: Default logic.
\newblock {\em Artificial Intelligence Review}, 8(1):3--16, 1994.

\bibitem[BED96]{bene-dech-96}
R.~Ben-Eliyahu and R.~Dechter.
\newblock Default reasoning using classical logic.
\newblock {\em Artificial Intelligence}, 84(1--2):113--150, 1996.

\bibitem[Bre91]{brew-91-b}
G.~Brewka.
\newblock Cumulative default logic: in defense of nonmonotonic inference rules.
\newblock {\em Artificial Intelligence}, 50(2):183--205, 1991.

\bibitem[BS90]{bopp-sips-90}
R.~Boppana and M.~Sipser.
\newblock The complexity of finite functions.
\newblock In J.~van Leeuwen, editor, {\em Handbook of Theoretical Computer
  Science}, volume~A, chapter~14, pages 757--804. Elsevier Science Publishers
  (North-Holland), Amsterdam, 1990.

\bibitem[CDLS00]{cado-etal-00}
M.~Cadoli, F.~M. Donini, P.~Liberatore, and M.~Schaerf.
\newblock Space efficiency of propositional knowledge representation
  formalisms.
\newblock {\em Journal of Artificial Intelligence Research}, 13:1--31, 2000.

\bibitem[CDLS02]{cado-etal-02}
M.~Cadoli, F.~Donini, P.~Liberatore, and M.~Schaerf.
\newblock Preprocessing of intractable problems.
\newblock {\em Information and Computation}, 176(2):89--120, 2002.

\bibitem[CMMT95]{chol-etal-95}
P.~Cholewinski, W.~Marek, A.~Mikitiuk, and M.~Truszczynski.
\newblock Experimenting with nonmonotonic reasoning.
\newblock In {\em Proceedings of the Twelfth International Conference on Logic
  Programming (ICLP'95)}, pages 267--281, 1995.

\bibitem[CS93]{cado-scha-93}
M.~Cadoli and M.~Schaerf.
\newblock A survey of complexity results for non-monotonic logics.
\newblock {\em Journal of Logic Programming}, 17:127--160, 1993.

\bibitem[DS03]{delg-scha-03}
J.~Delgrande and T.~Schaub.
\newblock On the relation between {R}eiter's default logic and its (major)
  variants.
\newblock In {\em Seventh European Conference on Symbolic and Quantitative
  Approaches to Reasoning with Uncertainty (ECSQARU 2003)}, pages 452--463,
  2003.

\bibitem[DSJ94]{delg-scha-jack-94}
J.~P. Delgrande, T.~Schaub, and W.~K. Jackson.
\newblock Alternative approaches to default logic.
\newblock {\em Artificial Intelligence}, 70:167--237, 1994.

\bibitem[FM92]{froi-meng-92}
C.~Froidevaux and J.~Mengin.
\newblock A framework for default logics.
\newblock In {\em European Workshop on Logics in AI (JELIA'92)}, pages
  154--173, 1992.

\bibitem[FM94]{froi-meng-94}
C.~Froidevaux and J.~Mengin.
\newblock Default logics: A unified view.
\newblock {\em Computational Intelligence}, 10:331--369, 1994.

\bibitem[GM94]{gior-mart-94}
L.~Giordano and A.~Martelli.
\newblock On cumulative default logics.
\newblock {\em Artificial Intelligence}, 66:161--179, 1994.

\bibitem[Got92]{gott-92-b}
G.~Gottlob.
\newblock Complexity results for nonmonotonic logics.
\newblock {\em Journal of Logic and Computation}, 2:397--425, 1992.

\bibitem[Got95]{gott-95}
G.~Gottlob.
\newblock Translating default logic into standard autoepistemic logic.
\newblock {\em Journal of the {ACM}}, 42:711--740, 1995.

\bibitem[Imi87]{imie-87}
T.~Imielinski.
\newblock Results on translating defaults to circumscription.
\newblock {\em Artificial Intelligence}, 32:131--146, 1987.

\bibitem[Jan98]{janh-98}
T.~Janhunen.
\newblock On the intertranslatability of autoepistemic, default and priority
  logics, and parallel circumscription.
\newblock In {\em Proceedings of the Sixth European Workshop on Logics in
  Artificial Intelligence (JELIA'98)}, pages 216--232, 1998.

\bibitem[Jan03]{janh-03}
T.~Janhunen.
\newblock Evaluating the effect of semi-normality on the expressiveness of
  defaults.
\newblock {\em Artificial Intelligence}, 144:233--250, 2003.

\bibitem[Kon88]{kono-88}
K.~Konolige.
\newblock On the relationship between default and autoepistemic logic.
\newblock {\em Artificial Intelligence}, 35:343--382, 1988.

\bibitem[Lib01]{libe-01}
P.~Liberatore.
\newblock Monotonic reductions, representative equivalence, and compilation of
  intractable problems.
\newblock {\em Journal of the {ACM}}, 48(6):1091--1125, 2001.

\bibitem[Lif94]{lifs-94}
V.~Lifschitz.
\newblock Circumscription.
\newblock In {\em Handbook of Logic in Artificial Intelligence and Logic
  Programming, Volume 3}, pages 297--352. Oxford University Press, 1994.

\bibitem[LLM03]{lang-etal-03}
J.~Lang, P.~Liberatore, and P.~Marquis.
\newblock Propositional independence: Formula-variable independence and
  forgetting.
\newblock {\em Journal of Artificial Intelligence Research}, 18:391--443, 2003.

\bibitem[Luk88]{luka-88}
W.~Lukaszewicz.
\newblock Considerations on default logic: an alternative approach.
\newblock {\em Computational Intelligence}, 4(1):1--16, 1988.

\bibitem[Moo85]{moor-85}
R.~C. Moore.
\newblock Semantical considerations on nonmonotonic logic.
\newblock {\em Artificial Intelligence}, 25:75--94, 1985.

\bibitem[MT95]{miki-trus-95}
A.~Mikitiuk and M.~Truszczynski.
\newblock Constrained and rational default logics.
\newblock In {\em Proceedings of the Fourteenth International Joint Conference
  on Artificial Intelligence (IJCAI'95)}, pages 1509--1517, 1995.

\bibitem[Pen00]{penn-00}
P.~Penna.
\newblock Succinct representations of model based belief revision.
\newblock In {\em Proceedings of the Sixteenth Symposium on Theoretical Aspects
  of Computer Science (STACS 2000)}, pages 205--216, 2000.

\bibitem[Rei80]{reit-80}
R.~Reiter.
\newblock A logic for default reasoning.
\newblock {\em Artificial Intelligence}, 13:81--132, 1980.

\bibitem[Ryc91]{rych-91}
P.~Rychlik.
\newblock Some variations on default logic.
\newblock In {\em Proceedings of the Ninth National Conference on Artificial
  Intelligence (AAAI'91)}, pages 373--378, 1991.

\bibitem[Sti92]{stil-92-b}
J.~Stillman.
\newblock The complexity of propositional default logics.
\newblock In {\em Proceedings of the Tenth National Conference on Artificial
  Intelligence (AAAI'92)}, pages 794--799, 1992.

\bibitem[Tur97]{turn-97}
H.~Turner.
\newblock Representing actions in logic programs and default theories: a
  situation calculus approach.
\newblock {\em Journal of Logic Programming}, 31(1--3):245--298, 1997.

\end{thebibliography}

\end{document}